\documentclass{ieeeaccess}
\usepackage{cite}
\usepackage{amsmath,amssymb,amsfonts}
\usepackage{algorithmic}
\usepackage{graphicx}
\usepackage{textcomp}

\usepackage{booktabs}
\usepackage{subcaption}
\usepackage{multirow}

\def\BibTeX{{\rm B\kern-.05em{\sc i\kern-.025em b}\kern-.08em
    T\kern-.1667em\lower.7ex\hbox{E}\kern-.125emX}}
\begin{document}
\history{Date of publication xxxx 00, 0000, date of current version xxxx 00, 0000.}
\doi{10.1109/ACCESS.2017.DOI}

\title{CroMe: Multimodal Fake News Detection using Cross-Modal Tri-Transformer and Metric Learning}
\author{
\uppercase{Eunjee Choi},
\uppercase{Junhyun Ahn},
\uppercase{XinYu Piao, and Jong-Kook Kim}, \IEEEmembership{Senior Member, IEEE}
}
\address{Department of Electrical and Computer Engineering, Korea University, Seoul 02841, Republic of Korea}
\tfootnote{This work was supported by the government of the Republic of Korea (MSIT) and the National Research Foundation of Korea (RS-2024-00466496), and supported by Korea Industrial Technology Association (KOITA) grant funded by Korea Ministry of Science and ICT (MSIT) (KOITA20250002-62)}

\markboth
{E. Choi \headeretal: CroMe: Multimodal Fake News Detection using Cross-Modal Tri-Transformer and Metric Learning}
{E. Choi \headeretal: CroMe: Multimodal Fake News Detection using Cross-Modal Tri-Transformer and Metric Learning}

\corresp{Corresponding author: XinYu Piao (e-mail: xypiao97@korea.ac.kr) and Jong-Kook Kim (e-mail: jongkook@korea.ac.kr).}

\begin{abstract}
Multimodal fake news detection has received increasing attention recently. Existing methods rely on independently encoded unimodal data and overlook the advantages of capturing intra-modal relationships and integrating inter-modal similarities using advanced techniques. To address these issues, Cross-Modal Tri-Transformer and Metric Learning (CroMe) for multimodal fake news detection is proposed. CroMe utilizes bootstrapping language-image pre-training (BLIP) with frozen image encoders and large language models as encoders to capture detailed text, image, and combined image-text representations. The metric learning module employs a proxy anchor method to capture intra-modality relationships while the feature fusion module uses a Cross-Modal and Tri-Transformer for effective integration. The final fake news detector processes the fused features through a classifier to predict the authenticity of the content. Experiments on datasets show that CroMe excels in multimodal fake news detection.
\end{abstract}


\titlepgskip=-15pt

\maketitle

\section{Introduction}\label{sec:introduction}
With rapid advances in information technology and the growing use of social media, digital online platforms have become the center of information exchange~\cite{mitra2017parsimonious}.
However, this has also led to a significant rise in fake news and misinformation, which may harm public opinion, disrupt political stability, and affect social and economic activities~\cite{zhu2022locating, shu2017fake, xia2023covid}.
Thus, effective false information detection has become essential to prevent these issues.
Traditional methods, such as identifying logical flaws or obvious signs like spelling errors and image alterations, were effective for text or image content separately.
However, the rise of multimedia formats like images and videos has accelerated the spread of fake news~\cite{jin2017multimodal}, making multimedia-focused detection methods necessary.

\begin{figure*}[t]
    \centering
    \vspace{0.1cm}
    \begin{subfigure}{0.212\textwidth}
        \centering
        \includegraphics[width=3.6cm,height=2.8cm]{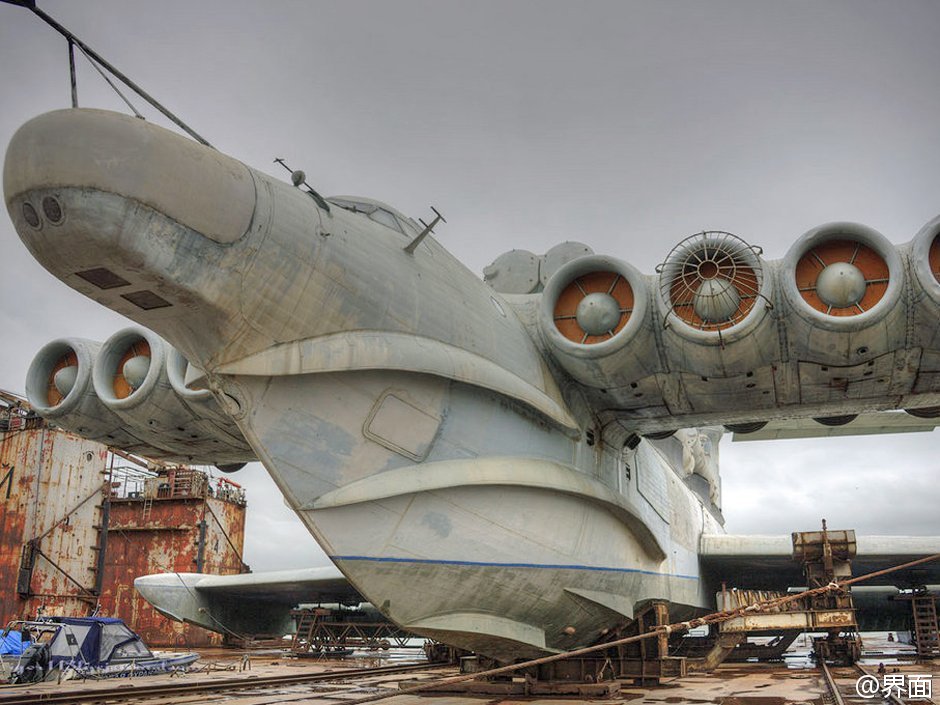} 
        \caption{A group of Quora users listed historically ingenious but unreasonable weapons.}
    \end{subfigure}
    \hspace{0.7cm}
    \begin{subfigure}{0.212\textwidth} 
        \centering
        \includegraphics[width=3.6cm,height=2.8cm]{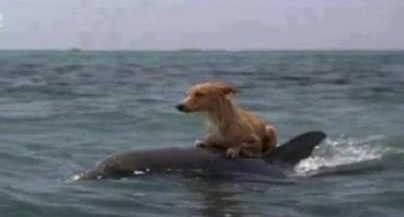}
        \caption{In Marco Island, dolphins rescued a dog and firefighters reunited it with its owner.}
    \end{subfigure}
    \hspace{0.7cm}
    \begin{subfigure}{0.212\textwidth} 
        \centering
        \includegraphics[width=3.6cm,height=2.8cm]{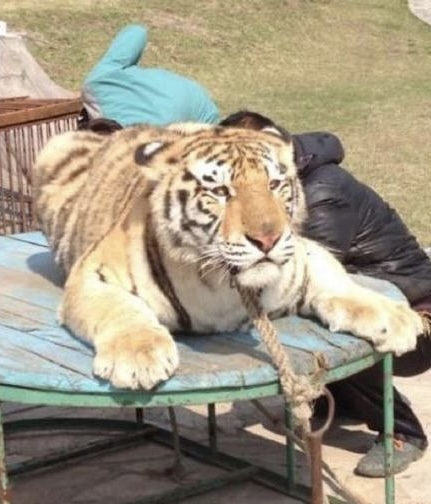}
        \caption{Yichang police seized 7 macaques and arrested 2 suspects.}
    \end{subfigure}
    \hspace{0.7cm}
    \begin{subfigure}{0.212\textwidth} 
        \centering
        \includegraphics[width=3.6cm,height=2.8cm]{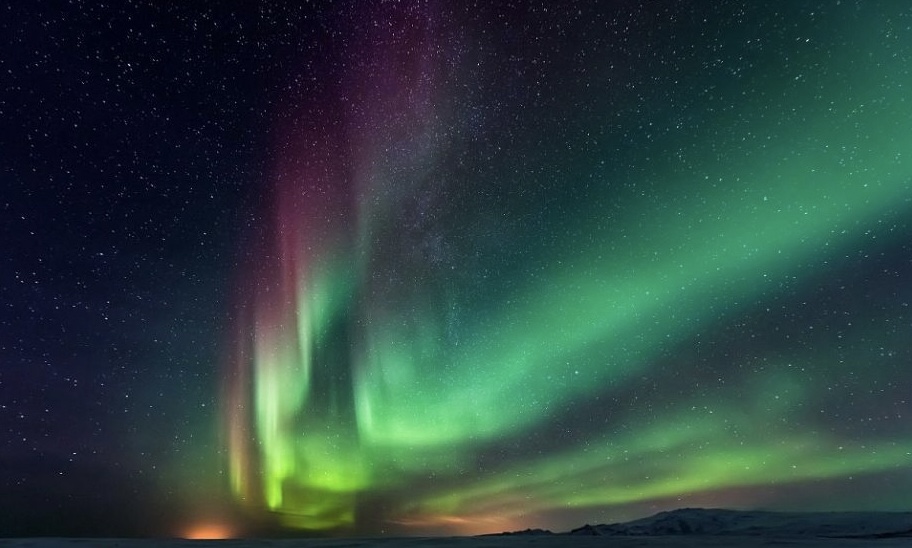}
        \caption{Photographer captured an abandoned Douglas DC 3 under Aurora's light.}
    \end{subfigure}
    \caption{Some fake examples from the Weibo and Weibo-21 datasets include: (a) The image shows intra-modality inconsistency. (b) Both the image and text contain intra-modality inconsistencies. (c) The image and text are unrelated, indicating inter-modality inconsistency. (d) The image does not match the text, reflecting inter-modality conflict.}
    \label{fig:intro}
\end{figure*}

Figure~\ref{fig:intro} shows four different examples of fake news from the Weibo~\cite{jin2017multimodal} and Weibo-21 datasets~\cite{nan2021mdfend}, which reveal various types of semantic inconsistencies.
The first example is fake news using a manipulated image, which shows intra-modal inconsistency.
Next, the second example shows that both the text and image are independently manipulated, demonstrating intra-modality inconsistencies within each modality.
In the third example, the tiger image is unrelated to the text, demonstrating inter-modality inconsistency.
The last example shows inter-modality conflict, where the text description does not fully align with the image.
Therefore, the main challenge in multimodal fake news detection is to effectively identify incongruent semantic features in both intra-modality and inter-modality contexts.
Recent advances in fake news detection have evolved from text-based methods to advanced deep learning approaches.
Early work focused on text analysis~\cite{castillo2011information}, while later studies employed deep neural networks (DNNs) for linguistic and temporal patterns~\cite{ma2016detecting}, and applied attention mechanisms with recurrent neural networks (RNNs)~\cite{chen2018call}. 
Jin et al.~\cite{jin2016novel} introduced multimodal approaches by combining image, text, and social context features. 
More recently, Choi et al.~\cite{10706486} developed TT-BLIP, integrating text, image, and multimodal features using BLIP~\cite{li2022blip} encoders and a MultiModal Tri-Transformer for feature fusion. 
Despite these advances, however, these approaches primarily focus on modality combination without addressing inter-modal and intra-modal relationships.

This paper introduces a multimodal fake news detection approach using the \underline{Cro}ss-Modal Tri-Transformer and \underline{Me}tric Learning (CroMe) network, a motivated framework inspired by the TT-BLIP model. 
To capture detailed text, image, and combined image-text features, CroMe uses pre-trained BERT~\cite{devlin-etal-2019-bert} for text encoding, Masked Autoencoders (MAE)~\cite{he2022masked} for image feature extraction, and BLIP2-OPT~\cite{li2023blip} for semantic features across modalities.
CroMe also introduces a \textit{Cross-Modal Fusion}~\cite{chen2022cross} to compute and integrate cross-modal similarities, addressing the limitations of existing methods in capturing complex inter-modal interactions. For intra-modal relationships, a type of metric learning called the \textit{Proxy Anchor Loss} method~\cite{schroff2015facenet,movshovitz2017no,kim2020proxy} is used. This method captures fine-grained similarities while maintaining the efficiency of proxy-based methods through dynamic gradients. Thus, CroMe ensures that features within the same modality are more closely aligned while capturing and integrating cross-modal correlations, enhancing the overall feature representation for more accurate fake news detection. Experiments performed using the Weibo and Weibo-21 datasets demonstrate that CroMe outperforms state-of-the-art models, highlighting its effectiveness in Multimodal Fake News Detection. For the Politifact dataset, CroMe achieved results comparable to state-of-the-art models because of the smaller dataset size.
The main contributions of this work are summarized as follows: 
\begin{itemize}
   \item The proposed multimodal fake news detection model, called CroMe, utilizes the advanced BLIP2-OPT model for feature extraction to capture detailed text, image, and combined image-text features.
   \item  The model uses a proxy anchor method in the metric learning module to capture intra-modality relationships ensuring effective representation learning within the same modality.
   \item The Cross-Modal Tri-Transformer Fusion, a new fusion technique, is introduced to calculate and merge similarities between text, image and image-text embeddings enhancing the model's ability to handle interactions among different modalities.
\end{itemize}
The paper is organized as follows.
Related work is described in Section~\ref{sec2}. 
Section~\ref{sec3} introduces the proposed CroMe model.
Then, Section~\ref{sec4} assesses its performance and Section~\ref{sec5} concludes the paper.

\begin{figure*}[!t]  
    \centering
    \includegraphics[width=0.985\textwidth]{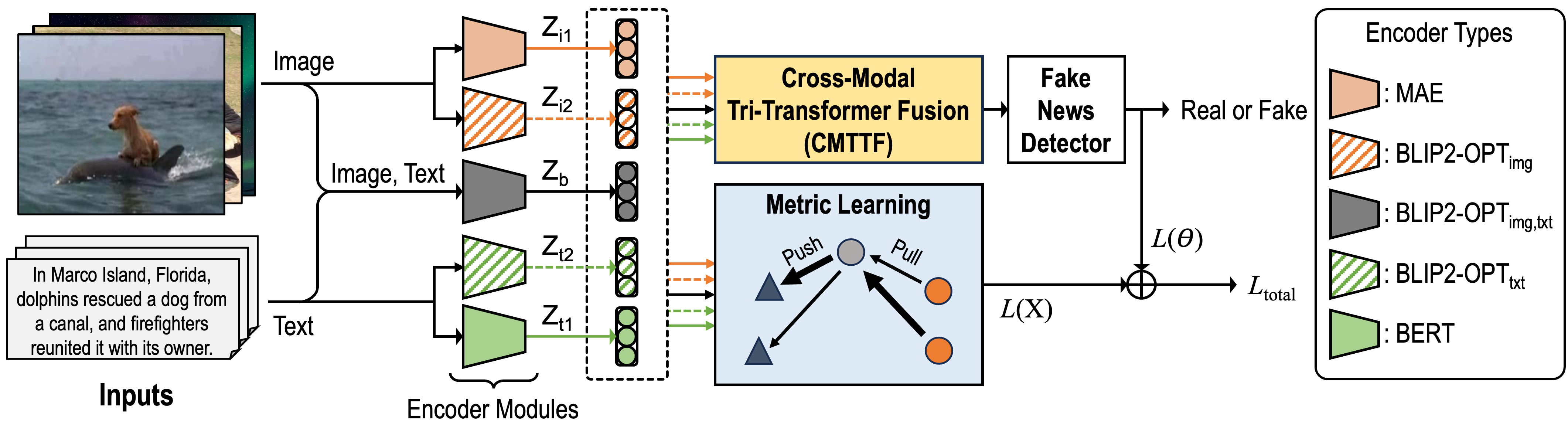}
    \caption{Overview of the CroMe architecture. Masked Autoencoder (MAE), BERT, and BLIP2-OPT encode multimodal features. Metric learning extracts intra-modal relationships by representing class data points with proxies, where arrow thickness indicates the gradient scale. CMTTF and fake news detector modules use cross-modal fusion and fake news detection.} 
    \label{fig:model_base}
\end{figure*}

\section{Related work}\label{sec2}
Several approaches have been developed for multimodal fake news detection, emphasizing the extraction of features from both images and text. EANN \cite{wang2018eann} enhances feature extraction using an event discriminator. MCAN \cite{wu2021multimodal} integrates textual and visual features through multiple co-attention layers. MVAE \cite{khattar2019mvae} employs a multimodal variational autoencoder to learn latent variables and reconstruct text and image features. Spotfake \cite{singhal2019spotfake} applies BERT \cite{devlin-etal-2019-bert} for text and VGG19 \cite{simonyan2014very} for image features, with Spotfake+ \cite{singhal2020spotfake+} extending this to full articles. SAFE \cite{zhou2020similarity} uses a similarity-based method to detect fake news by analyzing text and visual information, and CAFE \cite{chen2022cross} measures cross-modal ambiguity by calculating the Kullback-Leibler (KL) divergence \cite{kullback1951information} between unimodal feature distributions. LIIMR \cite{singhal2022leveraging} emphasizes the primary modality while minimizing the influence of less significant ones. DistilBert \cite{allein2021like} detects disinformation by correlating user preferences and sharing behaviors using latent representations of content. BDANN \cite{zhang2020bdann} combines features from two modalities to tackle event-specific biases in fake news detection on microblogs. FND-CLIP \cite{zhou2023multimodal} uses the CLIP vision-language model to measure image-text correlations, employing unimodal and pairwise CLIP encoders to aggregate features through modality-wise attention. 
TT-BLIP\cite{10706486} improves direction by using the BLIP\cite{li2022blip} model for unified vision-language understanding, with separate encoders for text, images, and multimodal data. 
The Multimodal Tri-Transformer then fuses these features using multi-head attention mechanisms for better data analysis.

Unlike previous models, the proposed model integrates unimodal features and cross-modal correlations. The metric learning module focuses on intra-modality relationships, ensuring that features from the same modality are closely aligned. This paper also captures inter-modality interactions by computing and integrating cross-modal similarities. By focusing on both intra-modality alignment and inter-modality integration, this approach improves fake news detection and outperforms the performance of state-of-the-art multimodal detection models using benchmark datasets.

\section{Methodology}
\label{sec3}

\begin{figure*}[t]
    \centering
    \begin{subfigure}{0.45\textwidth}
        \includegraphics[width=1\linewidth]{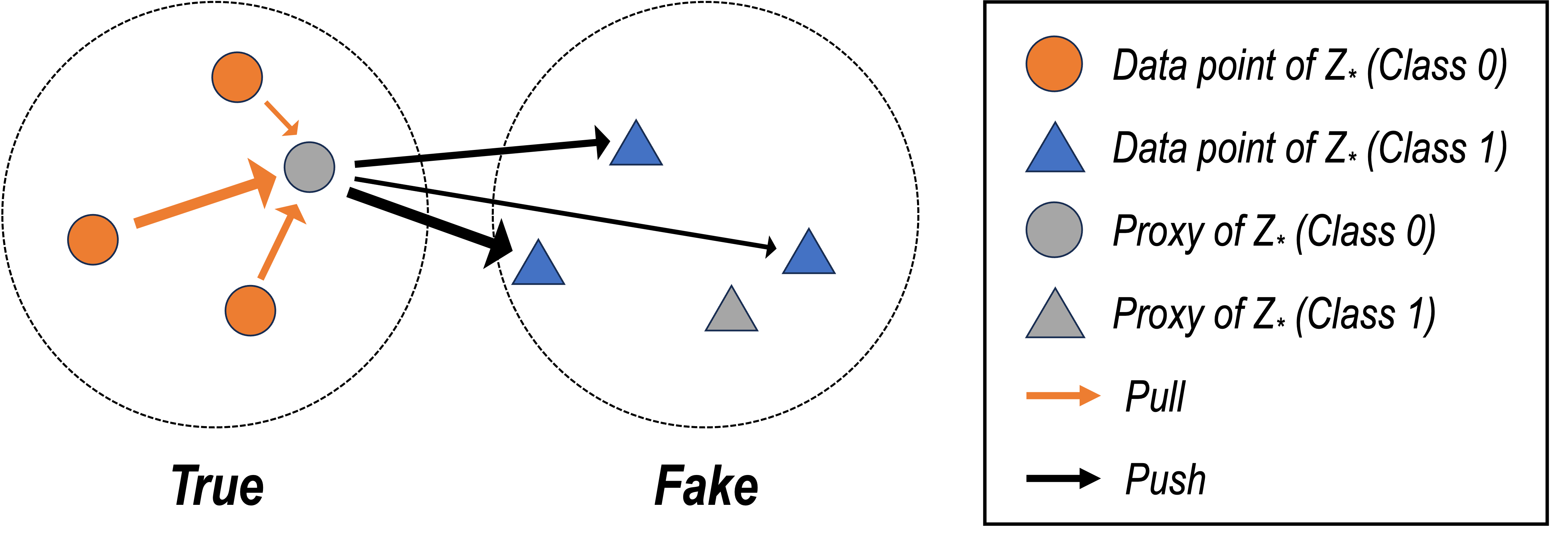} 
        \caption{Proxy anchor loss}
        \label{fig:pal_loss}
    \end{subfigure}
    \hspace{0.5cm}
    \begin{subfigure}{0.45\textwidth} 
        \includegraphics[width=1\linewidth]{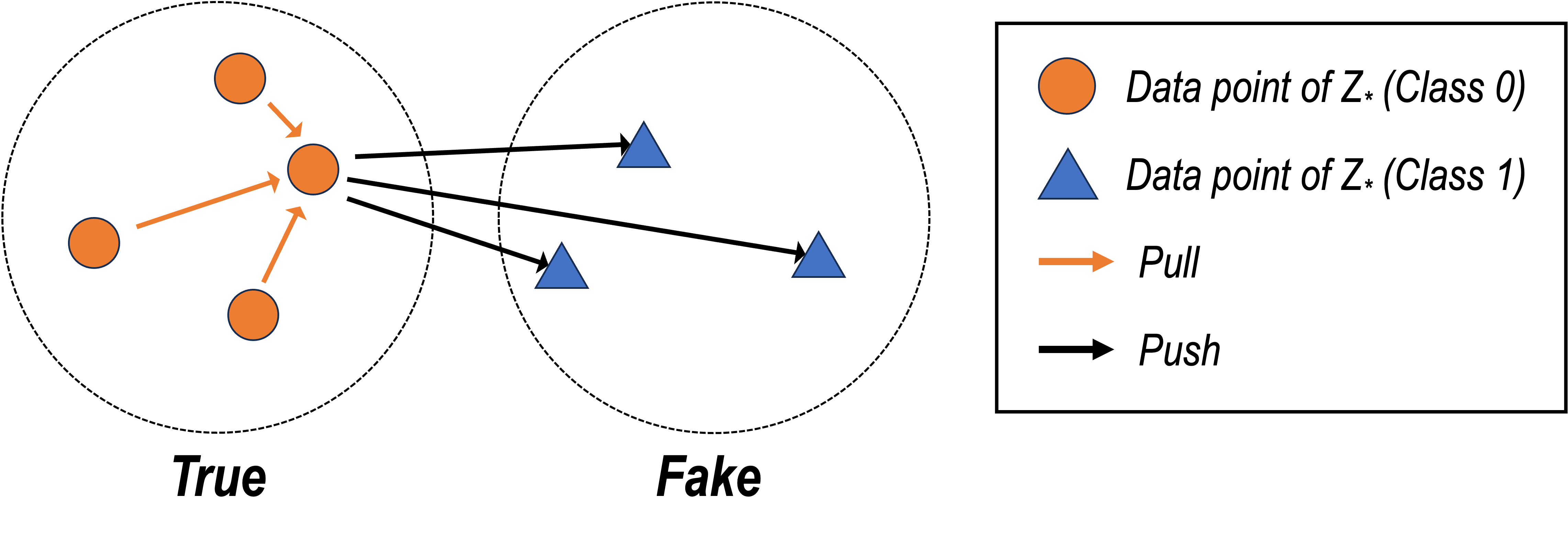}
        \caption{Triplet loss}
        \label{fig:triplet_loss}
    \end{subfigure}
    \caption{Comparison of two different metric learning methods; (a) Proxy as an anchor, and (b) Data point as an anchor. The thickness of arrows in proxy anchor loss indicates the gradient scale determined by the scaling factor $\alpha$.}
    \label{fig:metric_learning}
\end{figure*}

\begin{figure*}[t] 
    \centering
    \vspace{0.3cm}
    \begin{subfigure}[t]{0.49\textwidth}
        \centering
        \includegraphics[width=1\linewidth]{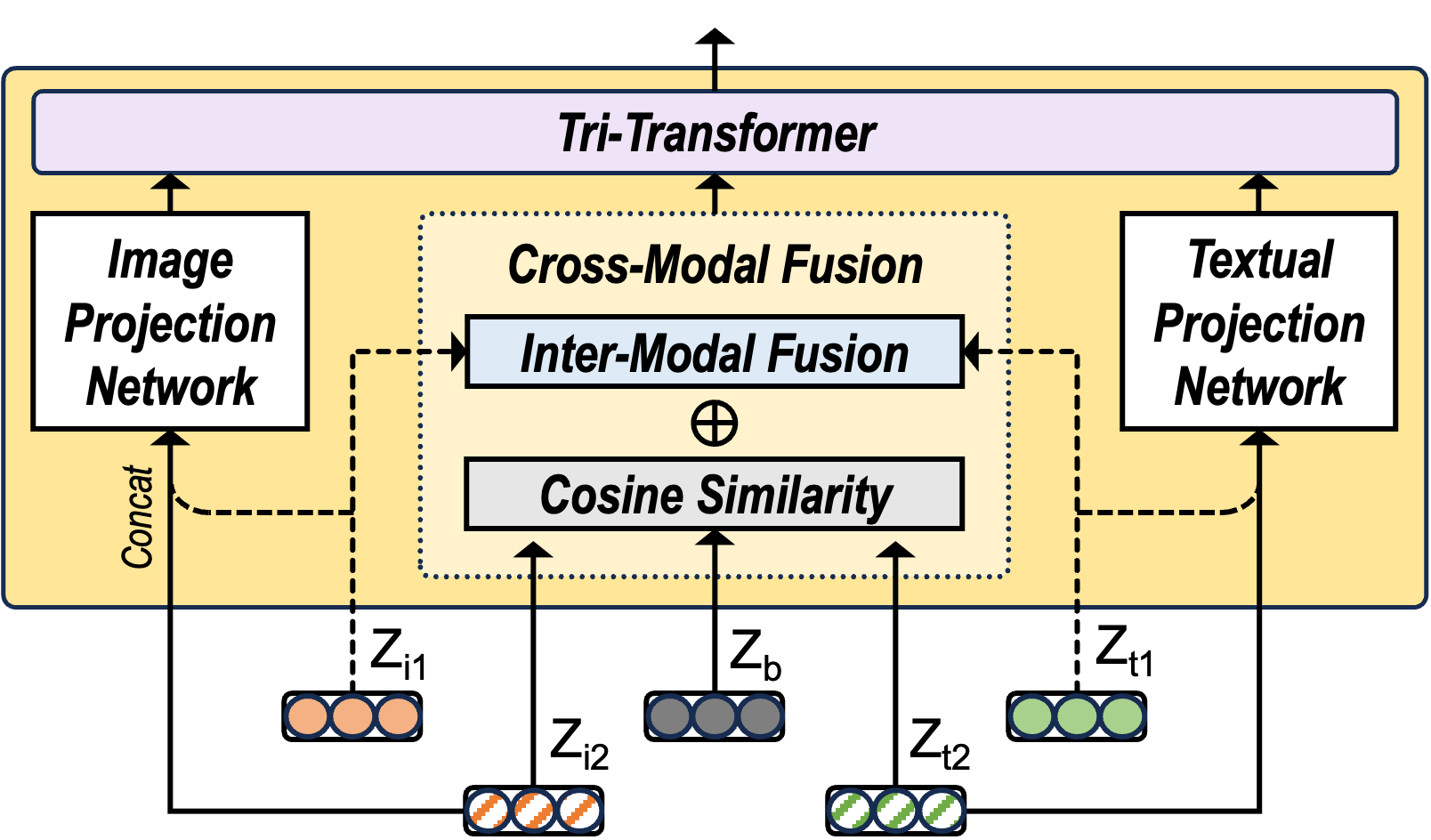} 
        \caption{CMTTF}
        \label{fig:cmttf_fusion}
    \end{subfigure}
    \begin{subfigure}[t]{0.245\textwidth}
        \centering
        \includegraphics[width=1\linewidth]{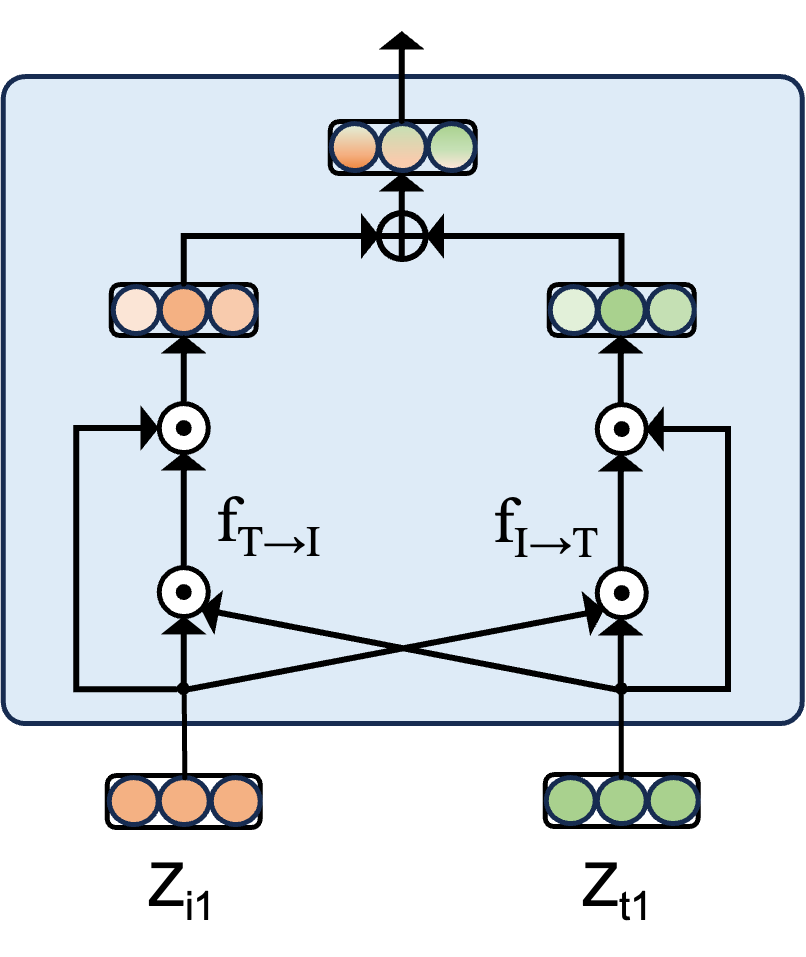}
        \caption{Inter-Modal Fusion}
        \label{fig:inter_modal_fusion}
    \end{subfigure}
    \begin{subfigure}[t]{0.245\textwidth}
        \centering
        \includegraphics[width=1\linewidth]{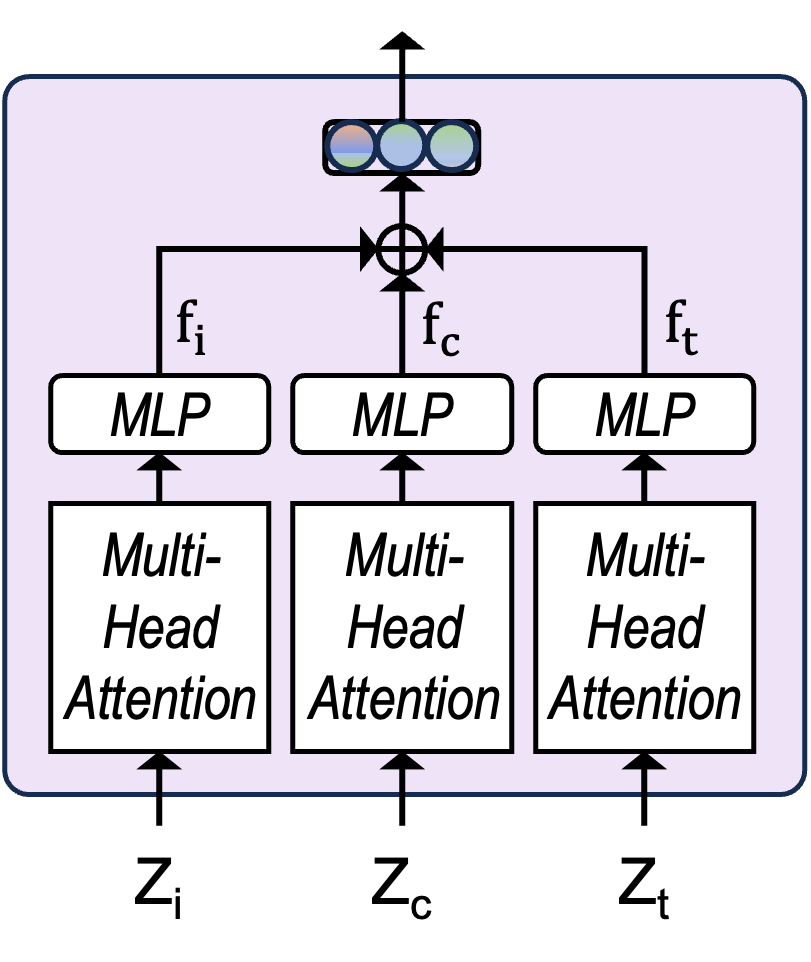}
        \caption{Tri-Transformer}
        \label{fig:tri_transformer}
    \end{subfigure}
    \caption{Overview of the CMTTF architecture, integrating Cross-Modal Fusion and Tri-Transformer modules to process and combine information from text, image, and image-text data. The architecture captures cross-modal correlations and fuses them to enhance feature representation.}
    \label{fig:CMTTF_Fusion}
\end{figure*}

\subsection{Overview}
CroMe, introduced in this paper and illustrated in Figure~\ref{fig:model_base}, is a multimodal model for fake news detection comprising four key modules. 
First, the encoder module extracts image encodings, text encodings, and integrated image-text encodings to capture modality-specific features from the input data.
Second, the metric learning module enhances intra-modality relationships using the proxy anchor method~\cite{kim2020proxy}, which effectively clusters same-class data points while separating different-class points. 
Third, the Cross-Modal Tri-Transformer Fusion (CMTTF) module fuses the modality-specific features generated by the encoder to capture complex cross-modal interactions and semantic relationships. 
Both the metric learning and CMTTF modules are applied in parallel to the encoder outputs, and their respective losses are combined to form the total loss $L_{\text{total}}$. This total loss is used to update the encoder parameters in the subsequent training epoch, enabling the encoder to capture both intra-modal and inter-modal relationships.
Lastly, the fake news detection module determines whether the news is real or fake from the fused representation.
This four-module architecture enables CroMe to effectively capture both modal-specific features and complex cross-modal interactions, resulting in more accurate fake news detection.

\subsection{Encoder Modules}
The encoder layer consists of components for encoding text, images and their combination. These components process input data denoted as \(x_{\text{img}}\) for images and \(x_{\text{txt}}\) for text.
For image data, the image encoder utilizes two parallel methods: Masked Autoencoders (MAE) \cite{he2022masked} and BLIP2-OPT Image. The pretrained MAE functions as the primary image encoder, producing encodings \(Z_{\text{i1}}=f_{\text{MAE}}(x_{\text{img}})\), capturing both global and local features by reconstructing missing data parts. In parallel, BLIP2-OPT Image provides an alternative encoding \(Z_{\text{i2}}=f_{\text{BLIP2-OPT}_{\text{img}}}(x_{\text{img}}, x_{\text{txt}})\), ensuring that image-specific encodings are derived without textual influence by utilizing a ``dummy text'' input. MAE focuses on global image structures, while BLIP2-OPT Image captures fine-grained semantics and multimodal associations, complementing each other.
For textual data, BERT and BLIP2-OPT Text encoders are utilized. The pretrained BERT \cite{devlin-etal-2019-bert} generates encodings \(Z_{\text{t1}}=f_{\text{BERT}}(x_{\text{txt}})\) by using its bidirectional understanding of text context. Additionally, BLIP2-OPT Text \cite{li2023blip} produces encodings \(Z_{\text{t2}} = f_{\text{BLIP2-OPT}_{\text{txt}}}(x_{\text{img}}, x_{\text{txt}})\) by focusing on text data alone, using a zero-filled image tensor to isolate the textual encoding process. While BERT excels in understanding text context, it lacks multimodal capabilities. In contrast, BLIP2-OPT Text, trained on multimodal datasets, is better suited for contexts. The outputs from these encoding processes are concatenated to form integrated image and text encodings, denoted as \(Z_{\text{i}}\) and \(Z_{\text{t}}\), respectively, for use in subsequent model components. 
The BLIP2-OPT model generates combined Image-Text encodings, capturing the cross-modal information between these modalities. This is represented as \(Z_{\text{b}} = f_{\text{BLIP2-OPT}_{\text{img,txt}}}(x_{\text{img}}, x_{\text{txt}} )\), incorporating the integrated features from both image and text inputs.

\subsection{Metric Learning}
Proxy anchor loss \cite{kim2020proxy} effectively learns data representations by utilizing distance relationships between data points and proxies. In this work, it is used to capture intra-modal relationships for different modalities, improving the distinction between data points in same class. The loss function \(L(X)\) uses static proxy assignment \cite{movshovitz2017no}, selecting a data point from each class as a proxy. It minimizes the distance between the proxy and same-class data points while maximizing the distance to those from different classes. Modality embeddings \(X\) are split into positive (\(X^+_p\)) and negative (\(X^-_p\)) sets relative to their proxies. The loss is defined as:
\begin{align}
    \begin{split}
        L(X) = & \frac{1}{|P^+|} \sum_{p \in P^+} \log \bigg( 1 + \sum_{x \in X^+_p} e^{-\alpha (s(x, p) - \delta)} \bigg) \\
               & +\frac{1}{|P|} \sum_{p \in P} \log \bigg( 1 + \sum_{x \in X^-_p} e^{\alpha (s(x, p) + \delta)} \bigg) \, ,
    \end{split}
    \label{eq:metric_learning_loss}
\end{align}
where \(\delta > 0\) is the margin parameter (margin) and \(\alpha > 0\) (alpha) is the scaling parameter.

As shown in Figure~\ref{fig:model_base}, the model iteratively trains each modality \( Z_{\text{*}} \) by fixing the parameters of other modalities, allowing it to refine intra-modal relationships without interference. This process is applied to each \( Z_{\text{*}} \in \{ Z_{\text{i1}}, Z_{\text{i2}}, Z_{\text{t1}}, Z_{\text{t2}}, Z_{\text{b}} \}\). Proxy Anchor Loss, illustrated in Figure~\ref{fig:pal_loss}, adjusts pull and push forces based on data point proximity, unlike traditional triplet loss~\cite{schroff2015facenet,peng2023mrml}. Proxy Anchor Loss is computationally efficient as it calculates distances between proxies and data points rather than between individual points, reducing computational load. Applied to text, image, and image-text features, it clusters data points of the same class while separating those of different classes, enhancing fake news detection.

\subsection{Cross-Modal Tri-Transformer Fusion}
This section introduces Cross-Modal Tri-Transformer Fusion (CMTTF) for integrating and processing image, text, and image-text data. CMTTF combines Cross-Modal Fusion~\cite{chen2022cross} and Tri-Transformer~\cite{10706486}, as illustrated in Figure~\ref{fig:cmttf_fusion}, with details on each component provided below.

\subsubsection{Cross-Modal Fusion}
To capture semantic interactions between modalities, the Cross-Modal Fusion module integrates text-image similarities using dot product with softmax normalization and cosine similarity~\cite{luo2018cosine}. Given unimodal representations \(Z_{\text{i1}}\), \(Z_{\text{t1}}\), and additional representations \(Z_{\text{i2}}\), \(Z_{\text{t2}}\), \(Z_{\text{b}}\), the process follows these steps: 1) \textit{Inter-modal Fusion} and 2) \textit{Cosine Similarity}.

First, \textit{Inter-modal Fusion} integrates semantic interactions between text and image modalities (is illustrated in Figure~\ref{fig:inter_modal_fusion}).
Correlations between text features \(\mathbf{t}\) and image features \(\mathbf{i}\) are computed, normalized, and combined into a unified representation \(\mathbf{C}_{\text{1}}\), as shown:
\begin{align}
    \mathbf{f}_{t \rightarrow i} & = \text{Softmax} \left( \frac{\mathbf{t} \cdot \mathbf{i}^\mathbf{T}}{\sqrt{d}} \right), \\[6pt]
    \mathbf{f}_{i \rightarrow t} & = \text{Softmax} \left( \frac{\mathbf{i} \cdot \mathbf{t}^\mathbf{T}}{\sqrt{d}} \right), \\[6pt]
    \mathbf{C}_{\text{1}} & = \mathbf{f}_{t \rightarrow i} \oplus \mathbf{f}_{i \rightarrow t} \, .
\end{align}
This inter-modal fusion approach suits models like BERT and MAE, processing unimodal data without embedding normalization. The dot product operation captures direct text-image interactions while preserving scale information, enabling effective multimodal representation.

Second, \textit{Cosine Similarity} is computed between different combinations of the inputs \(Z_{\text{i2}}\), \(Z_{\text{t2}}\), and \(Z_{\text{b}}\) to measure the relationships between the modalities. 
The similarities \(\mathcal{S}_{\text{ti}}\) (\(Z_{\text{t2}}\), \(Z_{\text{i2}}\)), \(\mathcal{S}_{\text{tb}}\) (\(Z_{\text{t2}}\), \(Z_{\text{b}}\)), and \(\mathcal{S}_{\text{ib}}\) (\(Z_{\text{i2}}\), \(Z_{\text{b}}\)) are computed using cosine similarity, defined for two vectors \(\mathbf{a}\) and \(\mathbf{b}\) as:
\begin{align}
    \mathcal{S}(\mathbf{a}, \mathbf{b}) = \frac{\mathbf{a} \cdot \mathbf{b}}{\|\mathbf{a}\| \times \|\mathbf{b}\|} \, .
\end{align}
The combined similarity is computed by weighting individual similarity scores with learnable parameters \(w_{\text{i2}}\), \(w_{\text{b}}\), \(w_{\text{t2}}\), and adding a bias term \(b\) to account for modality differences. This similarity is processed through linear layers, ReLU activations, and batch normalization, producing \(\mathbf{C}_{\text{2}}\). BLIP2-OPT components use cosine similarity as it effectively captures directional alignment, simplifies the fusion process, and maintains key correlations between modalities~\cite{li2023blip}.
The outputs \(\mathbf{C}_{\text{1}}\) and \(\mathbf{C}_{\text{2}}\) are combined into \(Z_{\text{c}}\), then processed through fully connected layers with batch normalization, ReLU activations, and dropout, resulting in the final correlation output used in the subsequent Tri-Transformer.

\subsubsection{Tri-Transformer}
The text and image embeddings (\(Z_{\text{i}}\) and \(Z_{\text{t}}\)) are passed through their respective projection networks and, along with correlation outputs (\(Z_{\text{c}}\)), are fed into the Tri-Transformer (as shown in Figure~\ref{fig:tri_transformer}).
This framework is based on the Multimodal Tri-Transformer as discussed in \cite{10706486}. 
Unlike TT-BLIP's text-focused cross-modal attention, this method processes text, image, and correlation independently.
\begin{align}
    f_{*} = \text{MultiHead}\left(\text{Softmax}\left(\frac{Q_* K_*^{\mathbf{T}}}{\sqrt{d_h}}\right) \times V_*; \theta_{\text{att}}^*\right)
\end{align}
Here, \(Q_*, K_*, V_*\) are query, key, and value matrices, \(\theta_{\text{att}}^*\) are attention parameters, and $* \in \{t, i, c\}$ corresponds to text, image, and correlation outputs respectively. 
Plus, \(d_h\) represents the dimensionality, with \(\mathbf{T}\) indicating the transpose operation. 
The outputs processed through Multi-Layer Perceptron (MLP) layers and combined into a unified tensor.

\begin{table*}[t]
    \centering
    \caption{Experimental results for Weibo, Weibo-21 and Politifact Datasets. A dash (‘-’) signifies that results are not provided in their corresponding research paper.}
    \begin{tabular}{llccccccc}  
        \toprule
        \multirow{2}{*}{\textbf{Datasets}} & \multirow{2}{*}{\textbf{Models}} & \multirow{2}{*}{\textbf{Accuracy}} & \multicolumn{3}{c}{\textbf{Fake News}} & \multicolumn{3}{c}{\textbf{Real News}} \\
        \cmidrule(lr){4-6} \cmidrule(lr){7-9}
        & & & Precision & Recall & F1 Score & Precision & Recall & F1 Score \\
        \midrule
        \multirow{12}{*}{Weibo} 
        & EANN \cite{wang2018eann} & 0.827 & 0.847 & 0.812 & 0.829 & 0.807 & 0.843 & 0.825 \\
        & MVAE \cite{khattar2019mvae} & 0.824 & 0.854 & 0.769 & 0.809 & 0.802 & 0.875 & 0.837 \\
        & MPFN \cite{jing2023multimodal} & 0.838 & 0.857 & 0.894 & 0.889 & 0.873 & 0.863 & 0.876\\
        & Spotfake \cite{singhal2019spotfake} & 0.892 & 0.902 & 0.964 & 0.932 & 0.847 & 0.656 & 0.739 \\
        & SAFE \cite{zhou2020similarity} & 0.762 & 0.831 & 0.724 & 0.774 & 0.695 & 0.811 & 0.748 \\
        & BDANN \cite{zhang2020bdann} & 0.821 & 0.790 & 0.610 & 0.690 & 0.830 & 0.920 & 0.870 \\
        & LIIMR \cite{singhal2022leveraging} & 0.900 & 0.882 & 0.823 & 0.847 & 0.908 & 0.941 & 0.925 \\
        & MCAN \cite{wu2021multimodal} & 0.899 & 0.913 & 0.889 & 0.901 & 0.884 & 0.909 & 0.897 \\
        & CAFE \cite{chen2022cross} & 0.840 & 0.855 & 0.830 & 0.842 & 0.825 & 0.851 & 0.837 \\
        & FND-CLIP \cite{zhou2023multimodal} & 0.907 & 0.914 & 0.901 & 0.908 & 0.914 & 0.901 & 0.907 \\
        & TT-BLIP \cite{10706486} & 0.961 & \textbf{0.979} & 0.944 & 0.961 & 0.944 & \textbf{0.980} & 0.962 \\ 
        & \textbf{CroMe (Ours)} & \textbf{0.974} & 0.964 & \textbf{0.984} & \textbf{0.974} & \textbf{0.985} & 0.966 & \textbf{0.975} \\ 
        \midrule
        \multirow{4}{*}{Weibo-21} 
        & EANN \cite{wang2018eann} & 0.870 & 0.902 & 0.825 & 0.862 & 0.841 & 0.912 & 0.875 \\
        & SpotFake \cite{singhal2019spotfake} & 0.851 & 0.953 & 0.733 & 0.828 & 0.786 & \textbf{0.964} & 0.866 \\
        & CAFE \cite{chen2022cross} & 0.882 & 0.857 & \textbf{0.915} & 0.885 & \textbf{0.907} & 0.844 & 0.876 \\
        & \textbf{CroMe (Ours)} & \textbf{0.917} & \textbf{0.944} & 0.917 & \textbf{0.930} & 0.880 & 0.918 & \textbf{0.899} \\

        \midrule
        \multirow{10}{*}{Politifact}
        & RoBERTa-MWSS \cite{shu2004leveraging} & 0.820 & - & - & 0.820 & - & - & - \\
        & SAFE \cite{zhou2020similarity} & 0.874 & 0.851 & 0.830 & 0.840 & 0.889 & 0.903 & 0.896 \\
        & Spotfake+ \cite{singhal2020spotfake+} & 0.846 & - & - & - & - & - & - \\
        & TM \cite{bhattarai-etal-2022-explainable} & 0.871 & - & - & - & 0.901 & - & - \\
        & LSTM-ATT \cite{lin2019detecting} & 0.832 & 0.828 & 0.832 & 0.830 & 0.836 & 0.832 & 0.829 \\
        & DistilBert \cite{allein2021like} & 0.741 & 0.875 & 0.636 & 0.737 & 0.647 & 0.880 & 0.746 \\
        & CAFE \cite{chen2022cross} & 0.864 & 0.724 & 0.778 & 0.750 & 0.895 & 0.919 & 0.907 \\
        & FND-CLIP \cite{zhou2023multimodal} & \textbf{0.942} & 0.897 & 0.897 & 0.897 & \textbf{0.960} & 0.960 & \textbf{0.960} \\
        & TT-BLIP \cite{10706486} & 0.904 & 0.913 & 0.724 & 0.808 & 0.901 & \textbf{0.973} & 0.936 \\
        & \textbf{CroMe (Ours)} & 0.933 & \textbf{0.987} & \textbf{0.925} & \textbf{0.955} & 0.793 & 0.958 & 0.868 \\
        \bottomrule
    \end{tabular}
    \label{tab:experimental_results_1}
\end{table*}

\begin{figure*}[t]
    \centering
    \begin{subfigure}{0.32\textwidth}
      \includegraphics[width=\textwidth]{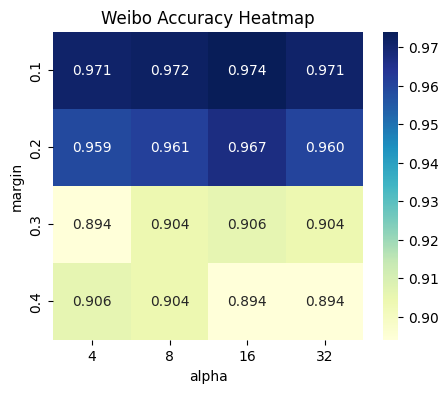}
      \caption{Weibo dataset}
      \label{fig:param_analysis_weibo}
    \end{subfigure}
    \begin{subfigure}{0.33\textwidth}
      \includegraphics[width=\textwidth]{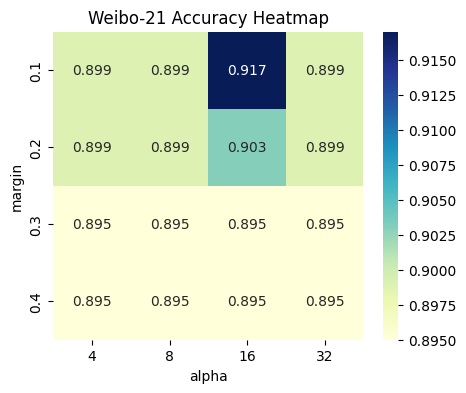}
      \caption{Weibo-21 dataset}
      \label{fig:param_analysis_weibo21}
    \end{subfigure}
    \begin{subfigure}{0.32\textwidth}
      \includegraphics[width=\textwidth]{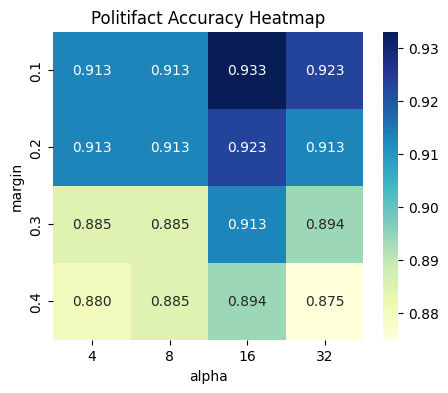}
      \caption{Politifact dataset}
      \label{fig:param_analysis_politifact}
    \end{subfigure}
    \caption{Parameter analysis for Weibo, Weibo-21 and Politifact dataset using the heatmap.}
    \label{fig:param_analysis}
\end{figure*}

\subsection{Fake News Detector}
Fake news detector uses the fusion representation to predict whether news articles are fake or real. This fused representation is processed by a classifier composed of three fully connected layers, ReLU activations, and batch normalization, resulting in a binary classification. The classifier assigns labels of `real' (0) or `fake' (1) to the news content, where \(\hat{y} = [\hat{y}_0, \hat{y}_1]\) denotes the probabilities of the news being `real' or `fake'. Cross-entropy is used to define the loss function $L(\theta)$, where $\theta$ represents the model parameters, as follows,
\begin{align}
    L(\theta) = -y\log(\hat{y}_1) - (1-y)\log(\hat{y}_0) \, .
\end{align}
The total loss $L_{\text{total}}$ is defined, where \(\beta\) controls the weight of this metric loss (as described in Equation~\ref{eq:metric_learning_loss}). The optimal value of \(\beta\) was determined through a grid search over the range [0.1, 1] using a step of 0.1. The model aims to minimize the total loss $L_{\text{total}}$ for each news data by learning $\theta$ through back-propagation.
\begin{align}
    L_{\text{total}} = L(\theta) +  \beta \cdot L(X)
\end{align}

\section{Evaluation}
\label{sec4}

\subsection{Experimental Setup}
\subsubsection{Datasets}
To evaluate the performance of CroMe, three datasets are used: Weibo~\cite{jin2017multimodal}, Weibo-21~\cite{nan2021mdfend}, and Politifact~\cite{shu2020fakenewsnet}. The Weibo dataset contains 6,137 training articles (2,802 fake and 3,335 real) and 1,685 test articles (833 fake and 852 real)~\cite{jin2017multimodal}. The Weibo-21 dataset includes 4,640 real and 4,487 fake articles, split into training and testing sets with an 8:2 ratio~\cite{nan2021mdfend}. The Politifact dataset consists of 381 training articles (246 fake and 135 real) and 104 test articles (74 fake and 30 real)~\cite{shu2020fakenewsnet}. Each dataset (Weibo, Weibo-21, and Politifact) was used separately for training and evaluation to avoid any data overlap or leakage.

\subsubsection{Training Settings}
Text encoding utilized the pretrained BERT model \cite{devlin-etal-2019-bert} for Chinese in the Weibo and Weibo-21 datasets and the "bert-base-uncased" model for the Politifact dataset. Images were resized to $224 \times 224$ pixels and encoded using Masked Autoencoders (MAE) \cite{he2022masked}. Image-text pairs were encoded with the pretrained BLIP2-OPT model \cite{li2023blip}, translating Chinese texts to English via the Google Translation API \cite{doi:10.1080/07317131.2012.650971}. The model employed five Adam optimizers for \( Z_{\text{*}} \in \{ Z_{\text{i1}}, Z_{\text{i2}}, Z_{\text{t1}}, Z_{\text{t2}}, Z_{\text{b}} \}\), with a learning rate of \(1 \times 10^{-3}\), a batch size of 64, and was trained for 50 epochs. Metric Learning parameters were set as follows: an iterative frequency of 5 epochs per modality, \(\alpha = 16\), \(\delta = 0.1\), and \(\beta = 0.1\), balancing classification and metric losses. These values were determined through preliminary experiments to optimize performance.
Experiments were performed five times, and the results are averaged across these runs.

\subsection{Results and Analysis}
CroMe’s performance is evaluated against state-of-the-art models shown in Table~\ref{tab:experimental_results_1}.
CroMe results were obtained from our implementation, while baseline results were cited from their respective original papers using the same datasets and evaluation metrics.
The evaluation metrics used include accuracy and precision, recall, and F1 scores for both real and fake news. CroMe achieved the highest accuracy of 0.974 in the experiments using Weibo, surpassing TT-BLIP by 1.3\% and FND-CLIP by 6.7\%. Similarly, when using Weibo-21, CroMe reached an accuracy of 0.917, outperforming CAFE by 3.5\% and EANN by 4.7\%. For both datasets, CroMe ranked 1st or 2nd in precision, recall, and F1 scores for both fake and real news. CroMe achieves 0.933 accuracy that closely matches FND-CLIP's 0.942. The reason may be that the dataset size is too small.

\begin{table*}[t]
    \centering
    \caption{Ablation experimental results of CroMe.}
    \begin{tabular}{lccccccccccc}
        \toprule
        \multirow{2}{*}{\textbf{Datasets}} & \multicolumn{8}{c}{\textbf{Modules}} & \multirow{2}{*}{\textbf{Accuracy}} & \multicolumn{2}{c}{\textbf{F1 Score}} \\
        \cmidrule(lr){2-9} \cmidrule(lr){11-12} & MAE & BLIP2-OPT\textsubscript{img} & BERT & BLIP2-OPT\textsubscript{txt} & BLIP2-OPT\textsubscript{img,txt} & CM & MT & TT & & Fake News & Real News \\
        \midrule
        \multirow{9}{*}{\centering Weibo} 
            &  &  & \checkmark & \checkmark & \checkmark & \checkmark & \checkmark & \checkmark & 0.961 & 0.919 & 0.975 \\
            & \checkmark & \checkmark & & & \checkmark & \checkmark & \checkmark & \checkmark & 0.950 & 0.852 & 0.961 \\
            & \checkmark & & \checkmark & & & \checkmark & \checkmark & \checkmark & 0.949 & 0.854 & 0.961 \\
            & \checkmark & \checkmark & \checkmark & \checkmark & & \checkmark & \checkmark  & \checkmark & 0.971 & 0.912 & \textbf{0.976} \\
            & \checkmark & \checkmark & \checkmark & \checkmark & \checkmark &  & \checkmark & \checkmark & 0.971 & 0.935 & 0.971 \\
            & \checkmark & \checkmark & \checkmark & \checkmark & \checkmark & \checkmark &  & \checkmark & 0.959 & 0.915 & 0.962 \\
            & \checkmark & \checkmark & \checkmark & \checkmark & \checkmark & \checkmark & \checkmark &   & 0.906 & 0.900 & 0.910\\ 

            & \checkmark & \checkmark & \checkmark & \checkmark & \checkmark &  & \checkmark  &  & 0.890 & 0.803 & 0.934\\
            
            & \checkmark & \checkmark & \checkmark & \checkmark & \checkmark & \checkmark & \checkmark & \checkmark  & \textbf{0.974} & \textbf{0.974} & 0.975\\ 
        \midrule
        \multirow{9}{*}{\centering Weibo21} 
            &  &  & \checkmark & \checkmark & \checkmark & \checkmark & \checkmark & \checkmark & 0.903 & 0.890 & 0.959 \\
            & \checkmark & \checkmark & & & \checkmark & \checkmark & \checkmark & \checkmark & 0.882 & 0.858 & 0.957 \\
            & \checkmark & & \checkmark & & & \checkmark & \checkmark & \checkmark & 0.890 & 0.876 & 0.957 \\
            & \checkmark & \checkmark & \checkmark & \checkmark & & \checkmark & \checkmark  & \checkmark & 0.897 & 0.870 & 0.916 \\  
            & \checkmark & \checkmark & \checkmark & \checkmark & \checkmark &  & \checkmark & \checkmark & 0.899 & 0.893 & \textbf{0.961} \\
            & \checkmark & \checkmark & \checkmark & \checkmark & \checkmark & \checkmark &  & \checkmark & 0.899 & 0.891 & 0.961 \\
            & \checkmark & \checkmark & \checkmark & \checkmark & \checkmark & \checkmark & \checkmark &  & 0.895 & 0.875 & 0.958\\ 
            & \checkmark & \checkmark & \checkmark & \checkmark & \checkmark &  & \checkmark  &  & 0.888 & 0.860 & 0.910\\ 
            
            & \checkmark & \checkmark & \checkmark & \checkmark & \checkmark & \checkmark & \checkmark & \checkmark  & \textbf{0.917} & \textbf{0.930} & 0.930\\
        \midrule
        \multirow{9}{*}{\centering Politifact} 
            &  &  & \checkmark & \checkmark & \checkmark & \checkmark & \checkmark & \checkmark & 0.913 & 0.927 & 0.972 \\
            & \checkmark & \checkmark & & & \checkmark & \checkmark & \checkmark & \checkmark & 0.865 & 0.891 & 0.968 \\
            & \checkmark & & \checkmark & & & \checkmark & \checkmark & \checkmark & 0.846 & 0.839 & 0.940 \\
            & \checkmark & \checkmark & \checkmark & \checkmark & & \checkmark & \checkmark  & \checkmark & 0.923 & 0.933 & 0.976 \\
            & \checkmark & \checkmark & \checkmark & \checkmark & \checkmark &  & \checkmark & \checkmark & 0.923 & 0.942 & \textbf{0.978} \\
            & \checkmark & \checkmark & \checkmark & \checkmark & \checkmark & \checkmark &  & \checkmark & 0.913 & 0.915 & 0.972 \\
            & \checkmark & \checkmark & \checkmark & \checkmark & \checkmark & \checkmark & \checkmark &  & 0.875 & 0.790 & 0.910\\ 
            & \checkmark & \checkmark & \checkmark & \checkmark & \checkmark &  & \checkmark  &  & 0.865 & 0.877 & 0.967\\         
            & \checkmark & \checkmark & \checkmark & \checkmark & \checkmark & \checkmark & \checkmark & \checkmark  & \textbf{0.933} & \textbf{0.955} & 0.868\\
        \bottomrule
    \end{tabular}
    \label{tab:experimental_results_2}
\end{table*}

\begin{figure*}[t]
  \centering
  \begin{subfigure}{0.16\linewidth}  
    \includegraphics[width=\linewidth]{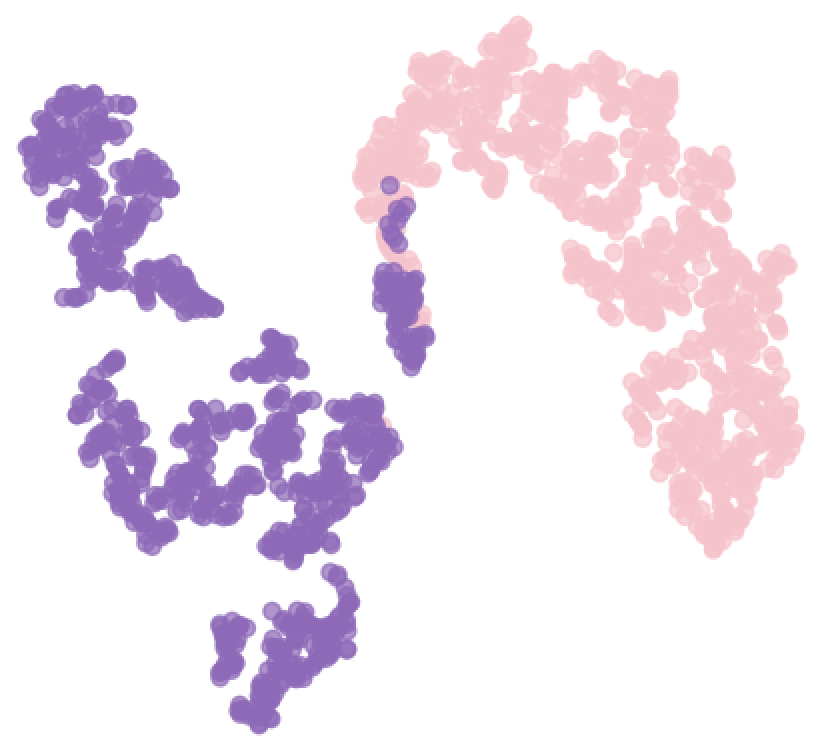}
    \caption{w/o image}
    \label{fig:image1}
  \end{subfigure}
  \begin{subfigure}{0.16\linewidth}  
    \includegraphics[width=\textwidth]{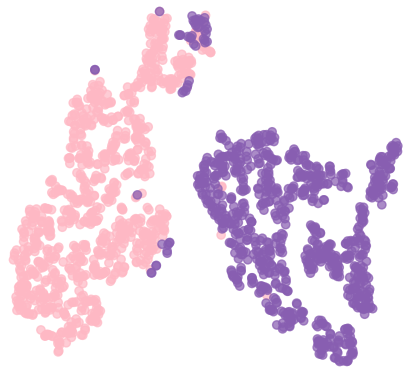}
    \caption{w/o text}
    \label{fig:image2}
  \end{subfigure}
  \begin{subfigure}{0.16\linewidth}  
    \includegraphics[width=\linewidth]{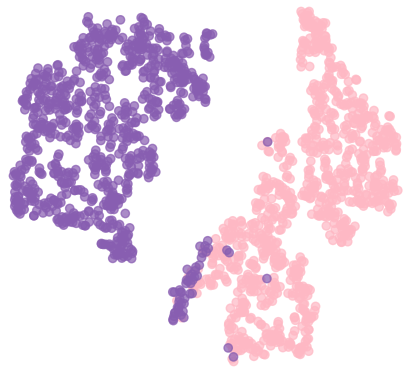}
    \caption{w/o BLIP2-OPT}
    \label{fig:image3}
  \end{subfigure}
  \begin{subfigure}{0.16\linewidth}  
    \includegraphics[width=\linewidth]{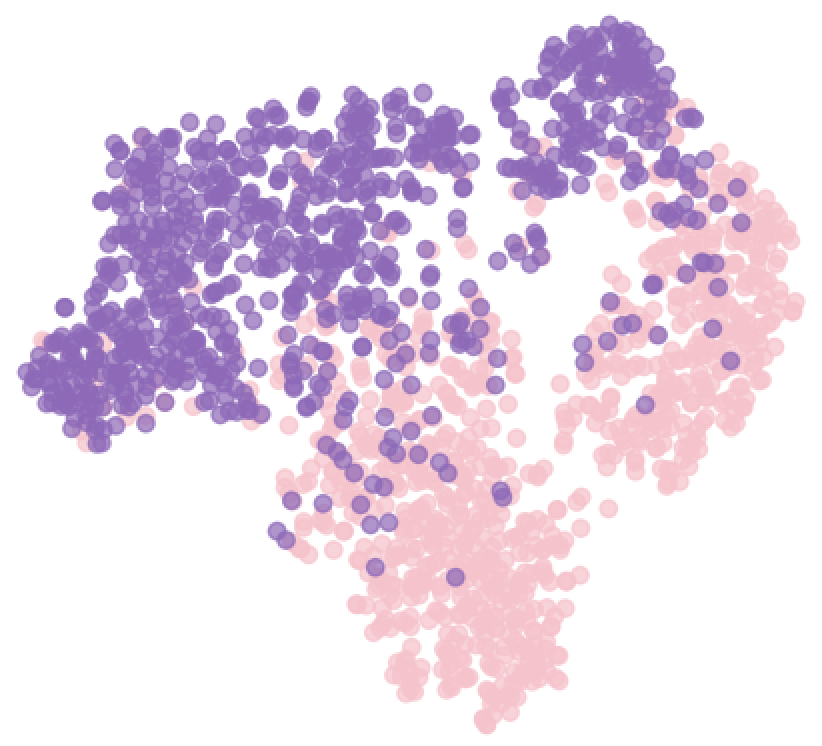}
    \caption{w/o CMTTF}
    \label{fig:image4}
  \end{subfigure}
  \begin{subfigure}{0.16\linewidth} 
    \includegraphics[width=\linewidth]{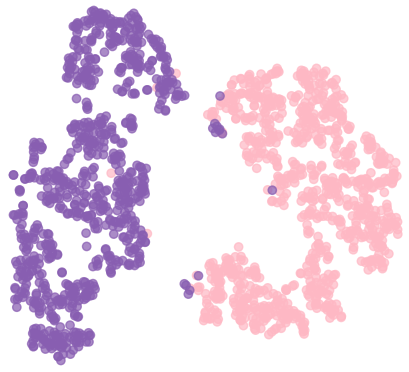}
    \caption{w/o MT}
    \label{fig:image5}
  \end{subfigure}
  \begin{subfigure}{0.16\linewidth} 
    \includegraphics[width=\linewidth]{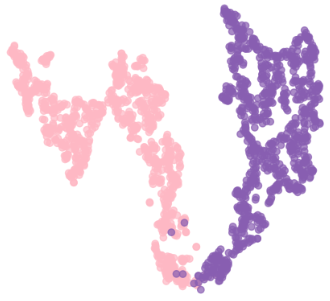}
    \caption{CroMe}
    \label{fig:image6}
  \end{subfigure}
  \caption{T-SNE visualizations of the features by CroMe and its variants using the Weibo test dataset, with each color representing a distinct label grouping.}
  \label{fig:t_sne}
\end{figure*}

CroMe excelled in fake news detection due to three key factors: 1) Advanced multimodal feature extraction through the BLIP2-OPT model enhances CroMe's capabilities by capturing detailed features from both text and images, allowing for more precise feature extraction. This precision improves the model's ability to distinguish between real and fake news. 2) The Cross-Modal Tri-Transformer Fusion (CMTTF) computes and integrates cross-modal similarities, enhancing the model's ability to utilize interactions between different modalities which is crucial for accurate fake news detection. 3) CroMe incorporates a Metric Learning module using the Proxy Anchor method, which focuses on intra-modality relationships. This ensures that features within the same modality are more closely aligned, leading to better representation learning and overall improved performance in detecting fake news. For smaller datasets, future research could explore data augmentation or transfer learning to improve performance.

The analysis of Weibo, Weibo-21, and Politifact datasets under different Proxy Anchor Loss hyperparameters, $\delta$ (margin) and $\alpha$ (alpha), is shown in Figure~\ref{fig:param_analysis}. As the margin increases, performance declines, with larger margins causing greater drops. Performance improves as alpha approaches 16, with the best results at alpha 16. Performance is more sensitive to margin variations than alpha. Alpha was tested at {4, 8, 16, 32}, and margin at {0.1, 0.2, 0.3, 0.4}. The best performance was achieved with alpha 16 and margin 0.1.

\subsection{Ablation}

Ablation experiments evaluated the impact of each component in the CroMe model. Table 2 shows the results for the Weibo, Weibo-21, and Politifact datasets.  The CroMe variants compared are:
\begin{enumerate}
    \item Without Image (BLIP2-OPT\textsubscript{img} + MAE): Removing the image encoders and using only text encoders and other components.
    \item Without Text (BLIP2-OPT\textsubscript{txt} + BERT): Removing the text encoders and using only image encoders and other components.
    \item Without BLIP2-OPT (BLIP2-OPT\textsubscript{img} + BLIP2-OPT\textsubscript{txt} + BLIP2-OPT\textsubscript{img,txt}): Removing all BLIP2-OPT components (text, image and image-text) and using only other encoders.
    \item Without BLIP2-OPT\textsubscript{img,txt}: Removing the image-text feature component of BLIP2-OPT, retaining the individual text and image BLIP2-OPT encoders.
    \item Without CM (Cross-Modal): Removing the Cross-Modal Fusion responsible for integrating cross-modal similarities.
    \item Without MT (Metric Learning): Removing the Metric Learning module.
    \item Without TT (Tri-Transformer): Removing the Tri-Transformer component.
\end{enumerate}
The ablation study highlights three components that most significantly impact accuracy when removed. First, excluding the BLIP2-OPT text components causes a substantial drop in accuracy across all datasets, as they are critical for understanding textual context necessary for identifying fake news. Second, removing the Cross-Modal Fusion (CM) component weakens performance by preventing effective integration of inter-modal fusion and cosine similarity features from text and images. Lastly, the Metric Learning module (MT) is essential for learning distinct features that differentiate fake and real news; its removal reduces accuracy. The ablation settings (Without Image, Without Text and Without CM) correspond respectively to fake news cases involving image tampering, text manipulation, and image–text mismatch, validating the capability of CroMe’s cross-modal fusion method. While the Tri-Transformer (TT) contributes to overall performance, the CM and MT modules distinguish CroMe from prior models such as TT-BLIP. CM improves cross-modal fusion, and MT enhances feature clustering within each modality. These modules result in improved classification performance, contributing to higher multimodal classification accuracy.

Figure~\ref{fig:t_sne} provides a T-SNE \cite{van2008visualizing} visualization of features before classification, comparing various CroMe settings: CroMe w/o image, CroMe w/o text, CroMe w/o BLIP, CroMe w/o fusion, CroMe w/o metric and the full CroMe model on the Weibo test dataset. Dots of the same color indicate the same label. 
Figure~\ref{fig:image4} shows that without the Cross-Modal Tri-Transformer Fusion (CMTTF) module, fake and real news instances are not well-separated. Including it improves clustering and distinction between classes. Comparisons of Figures~\ref{fig:image1},~\ref{fig:image2},~\ref{fig:image3}, and~\ref{fig:image6} shows that removing BLIP2-OPT feature extraction results in less clear clustering, highlighting its importance for integrating image, text, and image-text data. Excluding text features also reduces separation in t-SNE plots, showing that image and image-text features alone are less effective. Comparing CroMe w/o MT (Metric Learning, Figure~\ref{fig:image5}) and CroMe (Figure~\ref{fig:image6}), the degree of separation of the sample dots in Figure~\ref{fig:image6} is higher. This indicates that capturing intra-modality relationships ensures effective representation learning within the same modality.

\subsection{Discussion}\label{sec4.5}
CroMe Architecture relies on machine translation when encoding Chinese posts using BLIP2-OPT. This dependency can introduce translation artifacts and culture-specific semantic drift. In addition, the Politifact dataset split (381/104 train/test) is comparatively small, and our slightly lower accuracy on this dataset suggests sensitivity to limited data. Future work should validate CroMe on larger and more diverse corpora, include cross-dataset transfer tests, and assess robustness when modalities are missing at inference (text-only or image-only). Data augmentation or transfer learning may mitigate the small-data effect.

\section{Conclusion}\label{sec5}
This study introduces Cross-Modal Tri-Transformer and Metric Learning (CroMe), a Multimodal Fake News Detection model. CroMe uses Bootstrapping Language-Image Pre-training with Frozen Image Encoders and Large Language Models (BLIP2-OPT) to capture intra-modality and inter-modality relationships. The model includes four main modules: encoders (BERT, BLIP2-OPT-text, Masked Autoencoders (MAE), and BLIP2-OPT-image), a metric learning module (proxy anchor method), a feature fusion module (Cross-Modal Tri-Transformer Fusion, CMTTF), and a fake news detection module. CroMe uses BLIP2-OPT for semantic information extraction, CMTTF for feature fusion, and the proxy anchor method for metric learning. It improves accuracy by 1.3\% on the Weibo dataset and 3.5\% on the Weibo-21 dataset compared to previous models. CroMe performed slightly below the state-of-the-art model by 0.9\% in the case of Politifact, due to the smaller dataset size.

\bibliographystyle{IEEEtran}
\bibliography{refs}

\begin{IEEEbiography}[{\includegraphics[width=1in,height=1.25in,clip,keepaspectratio]{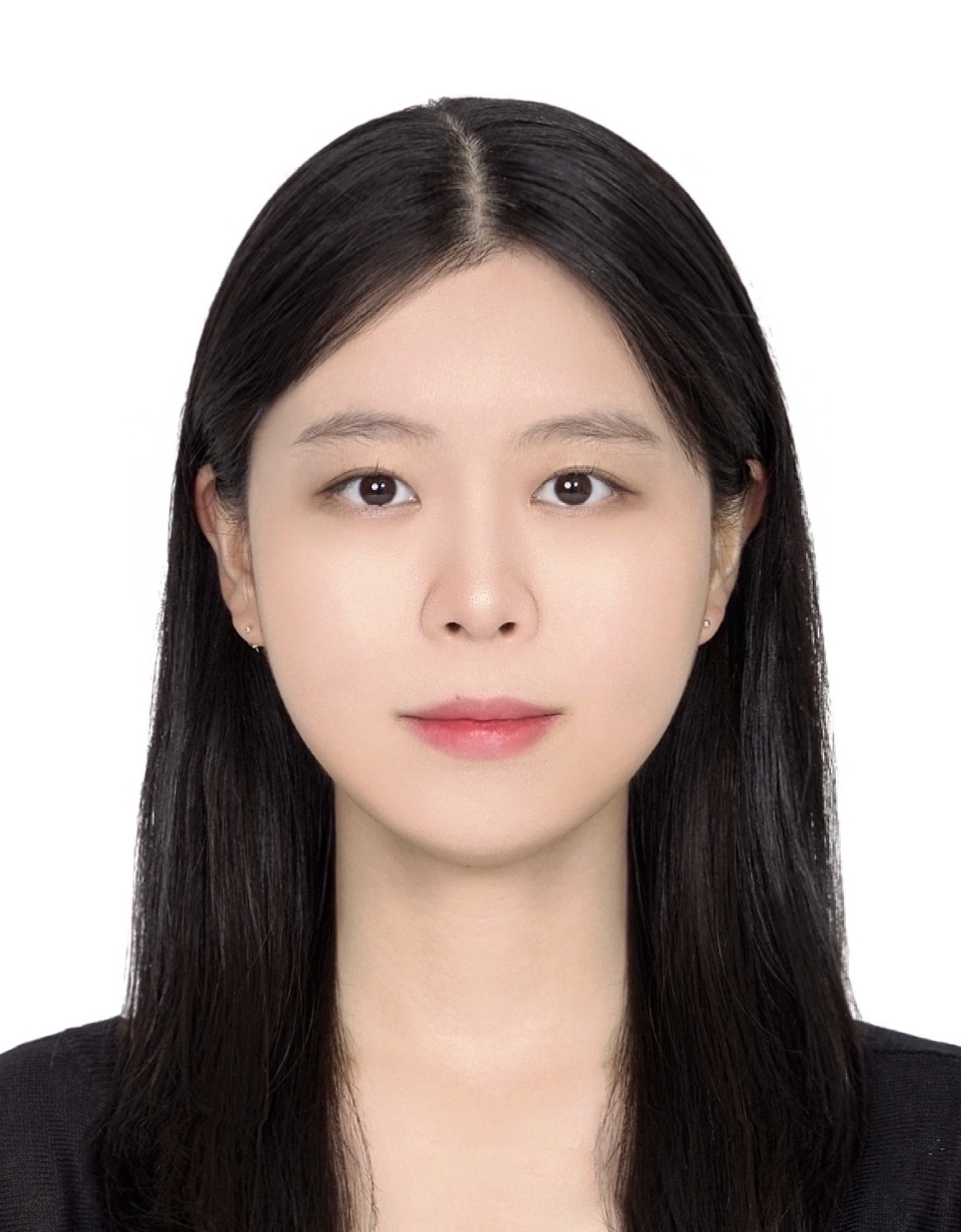}}]{EunJee Choi}
received the B.S. degree from the Department of Business Administration, Dankook University, in 2022, and the M.S. degree from the Department of Electrical and Computer Engineering, Korea University, Seoul, Republic of Korea, in 2025.
Her research interests include multimodal learning, fake news detection, large language models, and visual language models.
\end{IEEEbiography}

\begin{IEEEbiography}[{\includegraphics[width=1in,height=1.25in,clip,keepaspectratio]{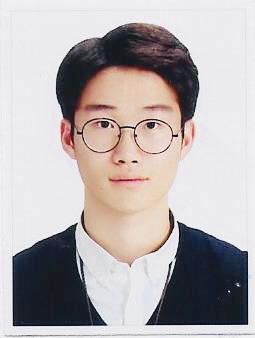}}]{Junhyun Ahn} received the B.S. degree in Electronic Engineering from Korea Aerospace University, Goyang, Republic of Korea, in 2024 and is currently pursuing the M.S. degree at the Department of Electrical and Computer Engineering, Korea University, Seoul, Republic of Korea.
His research interests include artificial neural networks and deep learning.
\end{IEEEbiography}

\begin{IEEEbiography}[{\includegraphics[width=1in,height=1.25in,clip,keepaspectratio]{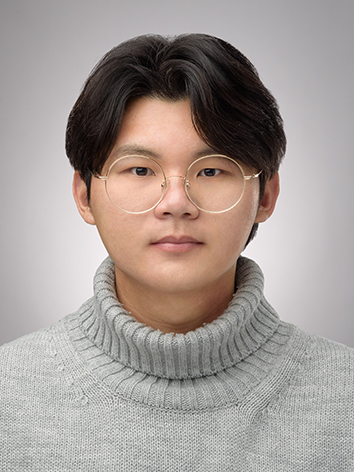}}]{XinYu Piao}
received the B.S. degree from the School of Electrical Engineering, Korea University, in 2020, and the Ph.D. degree from the Department of Electrical and Computer Engineering, Korea University, Seoul, Republic of Korea, in 2025. He is currently a postdoctoral researcher at the Department of Electrical and Computer Engineering, Korea University, since April 2025.

His research interests include systems for artificial intelligence (AI), hardware resource management, deep learning applications, and quantum computing.
\end{IEEEbiography}

\begin{IEEEbiography}[{\includegraphics[width=1in,height=1.25in,clip,keepaspectratio]{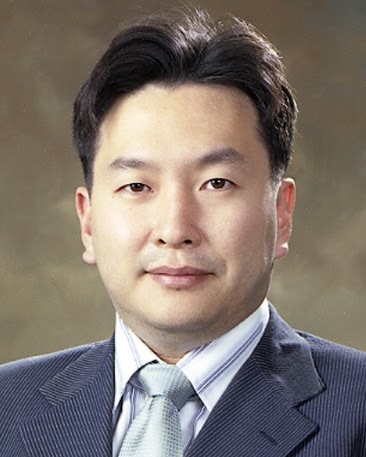}}]{Jong-Kook Kim} (Senior Member, IEEE)
received the B.S. degree in electronic engineering from Korea University, Seoul, South Korea, in 1998, and the M.S. and Ph.D. degrees from the School of Electrical and Computer Engineering, Purdue University, in 2000 and 2004, respectively. He is currently a Professor with the School of Electrical Engineering, Korea University, where he joined in 2007. 
He was with the Samsung SDS IT Research and Development Center from 2005 to 2007. 

His research interests include heterogeneous distributed computing, energy-aware computing, resource management, evolutionary heuristics, efficient systems and computing, artificial neural networks, deep learning, and systems for AI. He is a Senior Member of the IEEE and the ACM.
\end{IEEEbiography}

\EOD

\end{document}